\documentclass{article}

\PassOptionsToPackage{numbers, compress}{natbib}
 \usepackage[preprint]{neurips_2026}

\usepackage[utf8]{inputenc} 
\usepackage[T1]{fontenc}    
\usepackage{hyperref}       
\usepackage{url}            
\usepackage{booktabs}       
\usepackage{amsfonts}       
\usepackage{nicefrac}       
\usepackage{microtype}      
\usepackage{xcolor}         

\usepackage{multirow}

\usepackage{subfig}
\usepackage{amsmath,graphicx}
\usepackage{mathtools}
\usepackage{pifont}
\newcommand{\xmark}{\ding{55}} 
\usepackage{wrapfig, wrapfig}

\title{SEIS: Subspace-based Equivariance and Invariance Scores for Neural Representations}

\author{%
  Huahua Lin\quad Katayoun Farrahi\quad Xiaohao Cai\\
  University of Southampton\\
  \texttt{\{huahua.lin,k.farrahi,x.cai\}@soton.ac.uk} \\
}

\begin{document}

\maketitle

\begin{abstract}
Understanding how neural representations respond to geometric transformations is essential for evaluating whether learned features preserve meaningful spatial structure. Existing approaches primarily assess robustness primarily by comparing model outputs under transformed inputs, offering limited insight into how geometric information is organized within internal representations and failing to distinguish between information loss and re-encoding. In this work, we introduce SEIS (\textit{Subspace-based Equivariance and Invariance Scores}), a subspace metric for analyzing layer-wise feature representations under geometric transformations, disentangling equivariance from invariance without requiring labels or explicit knowledge of the transformation. Through controlled experiments across diverse architectures, we uncover several consistent patterns. First, convolutional encoders exhibit a depth-wise transition from strong equivariance to increasing invariance, with both properties stabilizing within the first few training epochs. In segmentation decoders, however, equivariance tends to recover in later layers. Second, this trade-off is not intrinsic but is shaped by training decisions: data augmentation actively strengthens both equivariance and invariance simultaneously, and multi-task learning induces synergistic gains in both properties beyond what either task achieves alone. Extending our analysis beyond convolutional networks, we find that transformer-based models exhibit distinct geometric behaviors, while MLP-Mixers display intermediate characteristics.
\end{abstract}

\section{Introduction}
\label{sec:intro}
The remarkable success of deep learning across computer vision \cite{he2016deep}, physics simulation \cite{sanchez2020learning}, and molecular biology \cite{jumper2021highly} is largely attributed to the effective utilization of inductive biases \cite{bronstein2017geometric}. Convolutional Neural Networks (CNNs) \cite{lecun1989backpropagation} achieve translation equivariance through weight sharing across spatial locations, while Group equivariant CNNs (G-CNNs) \cite{cohen2016group} extend this to rotation and reflection symmetries, and Graph Neural Networks (GNNs) \cite{weiler2019general} encode permutation symmetries as architectural constraints. However, many modern architectures, such as Vision Transformers (ViT) \cite{dosovitskiy2021image}, do not provide explicit geometric guarantees, and even architectures designed with such priors may exhibit behavior that deviates from idealized assumptions in practice \cite{azulay2019deep}. As a result, downstream performance alone provides an incomplete picture: a model may perform well while internally violating the geometric assumptions it was designed to respect \cite{geirhos2020shortcut, nguyen2015deep}.

The central challenge lies in quantifying these internal properties. Existing approaches \cite{goodfellow2009measuring, lenc2015understanding, kvinge2022ways, deng2022strong, bruintjes2023affects, quiroga2023invariance} typically evaluate whether model outputs remain stable under geometric transformations of the input. However, they require predefining specific target properties (e.g., translation equivariant) to evaluate. Moreover, such tests conflate two distinct failure modes: they cannot distinguish between representations that have lost geometric information entirely (functional failure) and those that preserve the information but re-encode it in a different basis (structural misalignment).

Representational similarity analysis compares neural representations directly without requiring predefined target properties. Canonical Correlation Analysis (CCA) \cite{hotelling1992relations} and its variants, including Singular Vector CCA (SVCCA) \cite{raghu2017svcca} and Centered Kernel Alignment (CKA) \cite{kornblith2019similarity}, are widely used to quantify similarity between neural layers by measuring correlations between their dominant subspaces across datasets or training stages. However, these tools are exclusively applied to cross-model or cross-layer comparisons, rather than analyzing the stability of a single representation under geometric transformation.

In this work, we address this gap by proposing a novel subspace-based metric/method \emph{Subspace-based Equivariance and Invariance Scores (SEIS)}, for analyzing how spatial structure is preserved under transformation. By reinterpreting subspace correlation analysis through a spatially-aware matricization, this approach allows us to explicitly disentangle two properties: \emph{equivariance}, defined as the preservation of information up to a linear transformation, and \emph{invariance}, defined as alignment of the spatial feature basis itself. This distinction enables us to identify when geometric information is preserved but re-encoded, a regime that is invisible to existing metrics.

Our main contributions and findings are as follows:
\begin{itemize}
    \item We introduce SEIS, the first subspace-based framework to explicitly disentangle equivariance from invariance within internal representations. By introducing a spatially-aware matricization strategy, SEIS enables layer-wise, label-free diagnosis of geometric stability without explicit transformation modeling.
    \item We show that the equivariance--invariance trade-off is not intrinsic but is systematically shaped by training dynamics and task objectives. Data augmentation actively strengthens both equivariance and invariance simultaneously rather than trading one for the other. Multi-task learning produces synergistic gains in both properties, with the jointly trained encoder surpassing single-task models on both metrics.
    \item We provide a unified empirical analysis across architectures, demonstrating that convolutional, transformer-based, and Multi-Layer Perceptron (MLP)-based models exhibit fundamentally different geometric behaviors, and identifying key architectural factors such as skip connections and positional encoding that govern the preservation of spatial structure.
\end{itemize}

\section{Related Work}
\label{sec:rw}
Quantifying the response of neural networks to geometric transformations is crucial for evaluating whether learned representations behave in a structured manner. Although specialized architectures such as G-CNNs \cite{cohen2016group} theoretically enforce equivariance, empirical measurement remains essential, as these guarantees often degrade in practice due to discretization effects on finite grids \cite{azulay2019deep} and mismatched symmetry assumptions, where imposed priors fail to align with the true data manifold \cite{benton2020learning}.

\noindent\textbf{Prediction Stability.} A common approach to measuring invariance is quantifying the stability of the final model predictions under input transformations. Metrics such as Accuracy Drop \cite{azulay2019deep} and Relative Corruption Errors \cite{hendrycks2019benchmarking} measure performance degradation under geometric shifts. To reduce dependence on ground-truth labels, Deng et al. \cite{deng2022strong} proposed Effective Invariance, which weighs stability by prediction confidence. Additionally, Softmax G-Empirical Equivariance Deviation (G-EED) \cite{kvinge2022ways} assesses global deviations using the Kullback–Leibler divergence between predictions and the group orbit average. While these methods provide an intuitive gauge of model invariance, they operate solely at the output level and treat the internal feature extraction as a black box. As a result, they cannot distinguish whether stability arises from robust semantic representations or shortcut learning via spurious correlations \cite{geirhos2020shortcut}.

\noindent\textbf{Element-Wise Feature Statistics.} To probe internal representations, significant effort has been directed toward analyzing the statistics of individual neurons or channels. Prior work has introduced statistical measures that compare a single neuron's response stability against its global firing profile \cite{goodfellow2009measuring, quiroga2023invariance}. More recently, equivariance has been quantified by computing cosine similarity and Pearson correlation between transformed channel activations and the channel activations of transformed inputs \cite{kvinge2022ways, bruintjes2023affects}. These methods typically aggregate scores by averaging across neurons or channels, implicitly assuming that units operate independently. However, deep neural networks rely on distributed representations where geometric information is encoded in the joint activation space rather than any single direction \cite{kornblith2019similarity}.

\noindent\textbf{Latent Transformation Modeling.} A more rigorous line of work seeks to model how transformations act within latent representations. Lenc et al. \cite{lenc2015understanding} measure equivariance by learning a linear mapping between representations of original and transformed inputs, and identify invariant features as those that are mapped approximately to themselves. However, invariance is treated as a special case of equivariance at the level of individual feature channels, rather than as a distinct property of the representation. As a result, this approach does not provide a unified or disentangled characterization of equivariance and invariance.


\noindent\textbf{Representation Similarity.} Subspace-based methods have been widely used to compare neural representations across models and layers by analyzing the eigenvectors or singular values of covariance or Gram matrices. CCA \cite{hotelling1992relations} and its variants, including SVCCA \cite{raghu2017svcca} and CKA \cite{kornblith2019similarity}, capture similarity up to invertible linear transformations, while Procrustes alignment \cite{ding2021grounding} enforces orthogonal correspondence. These methods have not been used to directly assess equivariance or invariance within a representation under input transformations. They treat spatial locations as independent samples when constructing similarity measures, thereby discarding the underlying grid structure. However, their reliance on intrinsic subspace structure rather than explicit transformation priors suggests that they could, in principle, be adapted to analyze geometric stability even when the form of the transformation is unknown.

\begin{table}[t]
\centering
\caption{
Comparison of representative approaches for measuring equivariance and invariance. Our method SEIS supports layer-wise analysis, disentangles equivariance from invariance, and operates on joint representations without requiring explicit transformation priors.
}
\vspace{4pt}
{\fontsize{8.5}{8}\selectfont
\setlength{\tabcolsep}{3pt}
\begin{tabular}{lcccc}
\toprule
\multirow{2}{*}{Method Category}
& \multirow{2}{*}{Layer-wise}
& \shortstack{Disentangles}
& \shortstack{w/o Requiring}
& \multirow{2}{*}{Joint Encoding} \\
 & 
& \shortstack{Equivariance \& Invariance}
& \shortstack{Transformation Priors}
&  \\
\midrule
Prediction stability
& \xmark
& \xmark
& \xmark
& \xmark \\
Element-wise feature statistics
& \checkmark
& \xmark
& \xmark
& \xmark \\
Latent transformation modeling
& \checkmark
& \xmark
& \xmark
& \xmark \\
Representation similarity
& \checkmark
& \xmark
& \checkmark
& \checkmark \\
\textbf{SEIS (Ours)}
& \checkmark
& \checkmark
& \checkmark
& \checkmark \\
\bottomrule
\end{tabular}
}
\label{tab:rw_comparison}
\end{table}

\section{Method}
\label{sec:method}

\noindent\textbf{Rationale.} While the theoretical definition of equivariance permits arbitrary non-linear transformations, spatial transformations on grid-structured data act as linear operators on the feature space, and convolutional architectures process features primarily through linear operations. While element-wise non-linearities introduce deviations, geometric structure is approximately preserved as a linear relationship between transformed representations \cite{lenc2015understanding}. Building on this insight, we introduce SEIS (equivariance and invariance scores $\mathcal{S}_\text{equiv}$ and $\mathcal{S}_\text{inv}$ in Section \ref{subsec:seis}), a subspace-based metric to diagnose how geometric information is preserved within neural representations under transformations, distinct from existing measures summarized in Table~\ref{tab:rw_comparison}.

\subsection{Notions of Equivariance and Invariance}
Let $f(\cdot)$ denote a fixed neural network layer. Given an input $x$ and its transformed counterpart $T(x)$, we denote the layer activations as $f(x)$ and $f(T(x))$, respectively. We formalize two notions of geometric consistency.

\noindent\textbf{Equivariance.} A representation is equivariant if the information encoded in $f(x)$ is preserved in $f(T(x))$ up to a linear change of basis. Formally, equivariance holds if there exists a linear operator $\mathbf{M}$ such that $f(T(x))\approx\mathbf{M}f(x)$. This implies functional preservation of information, allowing features to be redistributed (e.g., permuted) while remaining linearly recoverable.

\noindent\textbf{Invariance.} Invariance imposes a stronger constraint: the feature basis itself must remain aligned. This implies that the mapping $\mathbf{M}$ is approximately the identity (up to sign or permutation), i.e., $f(T(x))\approx f(x)$. This reflects robustness at the level of feature organization, where the representation responds consistently without requiring a change of basis.

\subsection{Spatially-Aware Matricization}
Let $\mathbf{Z} \in \mathbb{R}^{b \times c \times h \times w}$ denote the activation tensor for a batch of inputs, where $b$, $c$, $h$, and $w$ correspond to batch size, channel count, and spatial dimensions. Given a spatial transformation $T$ applied to the input $x$, we obtain paired activations $\mathbf{Z}=f(x)$ and $\mathbf{Z}'=f(T(x))$.

Unlike standard representational similarity (e.g., CCA and CKA), which flatten spatial dimensions into the sample axis and compare channel-wise feature subspaces, SEIS treats spatial locations as features and aggregates channel responses across samples as observations. Specifically, we reshape the activations into matrices $\mathbf{A}, \mathbf{A}' \in \mathbb{R}^{d \times n}$, where $d = h \cdot w$ indexes spatial coordinates and $n = b \cdot c$ indexes observations. This inversion allows us to measure the stability of the spatial feature basis itself, rather than the alignment of channel-wise feature subspaces.

Under our representation, a spatial transformation acts as a permutation or interpolation operator on the feature axis. Consequently, linear relationships between $\mathbf{A}$ and $\mathbf{A}'$ directly reflect whether spatial information is preserved, redistributed, or destroyed by the transformation.

\subsection{Subspace Denoising via SVD}
\label{sec:3-3}
Prior work has shown that the effective rank of neural representations is significantly lower than their ambient dimensionality, especially in deeper layers \cite{ansuini2019intrinsic, gong2019intrinsic}. To focus the analysis on the principal subspace, we perform Singular Value Decomposition (SVD) on each matrix,
\begin{equation}
    \mathbf{A} = \mathbf{U}_A \mathbf{\Sigma}_A \mathbf{V}_A^\top, \quad \mathbf{A}' = \mathbf{U}_{A'} \mathbf{\Sigma}_{A'} \mathbf{V}_{A'}^\top.
\end{equation}
We retain the leading $k_A$ and $k_{A'}$ left singular vectors $\tilde{\mathbf{U}}_A \in \mathbb{R}^{d \times k_{A}}$ and $\tilde{\mathbf{U}}_{A'} \in \mathbb{R}^{d \times k_{A'}}$ that explain $99\%$ of the cumulative variance, following the standard practice in SVCCA \cite{raghu2017svcca} (see Appendix~\ref{sec:appendix_vt_sensitivity} for a sensitivity analysis of this threshold), to obtain denoised subspace representations
\begin{equation}
\tilde{\mathbf{A}} = \tilde{\mathbf{U}}_A^\top\mathbf{A}\in\mathbb{R}^{k_A\times n}, \quad \tilde{\mathbf{A}}' = \tilde{\mathbf{U}}_{A'}^\top \mathbf{A}'\in\mathbb{R}^{k_{A'}\times n}.
\end{equation}
This step is critical for numerical stability, as applying CCA directly to high-dimensional activations often leads to ill-conditioned covariance estimates, resulting in numerically unstable and unreliable canonical directions \cite{won2013condition, morcos2018insights}.

\subsection{Quantifying Equivariance and Invariance}
\label{subsec:seis}
We apply CCA to the denoised subspaces $\tilde{\mathbf{A}}$ and $\tilde{\mathbf{A}}'$. CCA identifies pairs of projection vectors $\mathbf{w}_i\in \mathbb{R}^{k_A}$ and $\mathbf{v}_i \in \mathbb{R}^{k_{A'}}$ that maximize the correlation between projected variates
\begin{equation}
\mathbf{p}_i = \mathbf{w}_i^\top \tilde{\mathbf{A}} \in\mathbb{R}^n, \ \ 
\mathbf{q}_i = \mathbf{v}_i^\top \tilde{\mathbf{A}}'\in\mathbb{R}^n, \ \ i=1,\dots,r,
\end{equation}
where $r=\min(k_A,k_{A'})$. These canonical components capture directions of maximal shared information between the original and transformed representations.

\noindent\textbf{Equivariance Score.}
Functional equivariance requires that information to be preserved under transformation up to a linear change of basis. We quantify this by mean absolute cosine similarity between the canonical variates
\begin{equation}
\mathcal{S}_\text{equiv} = \frac{1}{r} \sum_{i=1}^{r} \left| \frac{\langle \mathbf{p}_i, \mathbf{q}_i \rangle}{\| \mathbf{p}_i \|_2 \| \mathbf{q}_i \|_2} \right|.
\end{equation}
Since canonical variates $\mathbf{p}_i$ and $\mathbf{q}_i$ are centered, this quantity is equivalent to the mean canonical correlation. A high $\mathcal{S}_{\text{equiv}}$ indicates that spatial information is preserved in the representation, even if redistributed across feature dimensions, consistent with linear equivariance.

\noindent\textbf{Invariance Score.}
Invariance imposes a stronger requirement: not only must information be preserved, but the spatial feature basis itself should remain aligned. To capture this, we measure the cosine similarity between the corresponding CCA projection vectors,
\begin{equation}
\mathcal{S}_\text{inv} = \frac{1}{r} \sum_{i=1}^{r} \rho_i \cdot \left| \frac{\langle \mathbf{w}_{i}, \mathbf{v}_{i} \rangle}{\| \mathbf{w}_{i} \|_2 \| \mathbf{v}_{i} \|_2} \right|,
\end{equation}
where $\rho_i$ denotes \textit{the $i$-th canonical correlation coefficient}. Weighting by $\rho_i$ suppresses directions that carry little shared information and ensures that alignment is assessed only where equivariance is meaningful. A high $\mathcal{S}_\text{inv}$ indicates that the features are not only recoverable but are spatially aligned, reflecting robustness to the transformation.

\section{Experiments}
\label{sec:exps}
We structure our evaluation to isolate how equivariance and invariance arise and interact across models. We begin with controlled experiments to validate the metric under known geometric transformations, and then analyze depth-wise behavior in convolutional networks to study how these properties evolve across layers. We further hypothesize that this behavior is shaped by training choices, including data augmentation and task objectives, which can strengthen or rebalance these properties. Finally, we extend the analysis across architectures, comparing convolutional, transformer-based, and MLP-based models, and examining the role of architectural components such as skip connections and positional encoding. All experiments are conducted on two NVIDIA A100 GPUs.

\subsection{Validation on Synthetic Transformations}
\noindent\textbf{Experimental Setup.} We validate SEIS using a controlled synthetic setup that isolates geometric effects from training dynamics. Activations are extracted from a single convolutional layer on MNIST, and paired representations are constructed by directly applying spatial transformations to the reference activations. This yields two ground-truth regimes: \emph{perfect invariance}, where the representation is unchanged, and \emph{strict equivariance}, where spatial transformations act linearly on the representation. This setting allows us to directly verify whether SEIS recovers known geometric relationships.

We evaluate six conditions: an identity control, four geometric transformations (translation, scaling, rotation, and a composite affine transform), and a random baseline. Transformation parameters are sampled stochastically: translation is limited to $\pm15\%$ of the spatial dimension; scaling factors range from $0.8$ to $1.2$; rotation is sampled from the full range of $360^\circ$; and a composite affine transformation combines all three distortions simultaneously. For each condition, both $\mathcal{S}_\text{equiv}$ and $\mathcal{S}_\text{inv}$ are computed and reported as the mean over 50 independent trials.

\begin{figure}[t]
    \centering
    \includegraphics[width=0.7\linewidth]{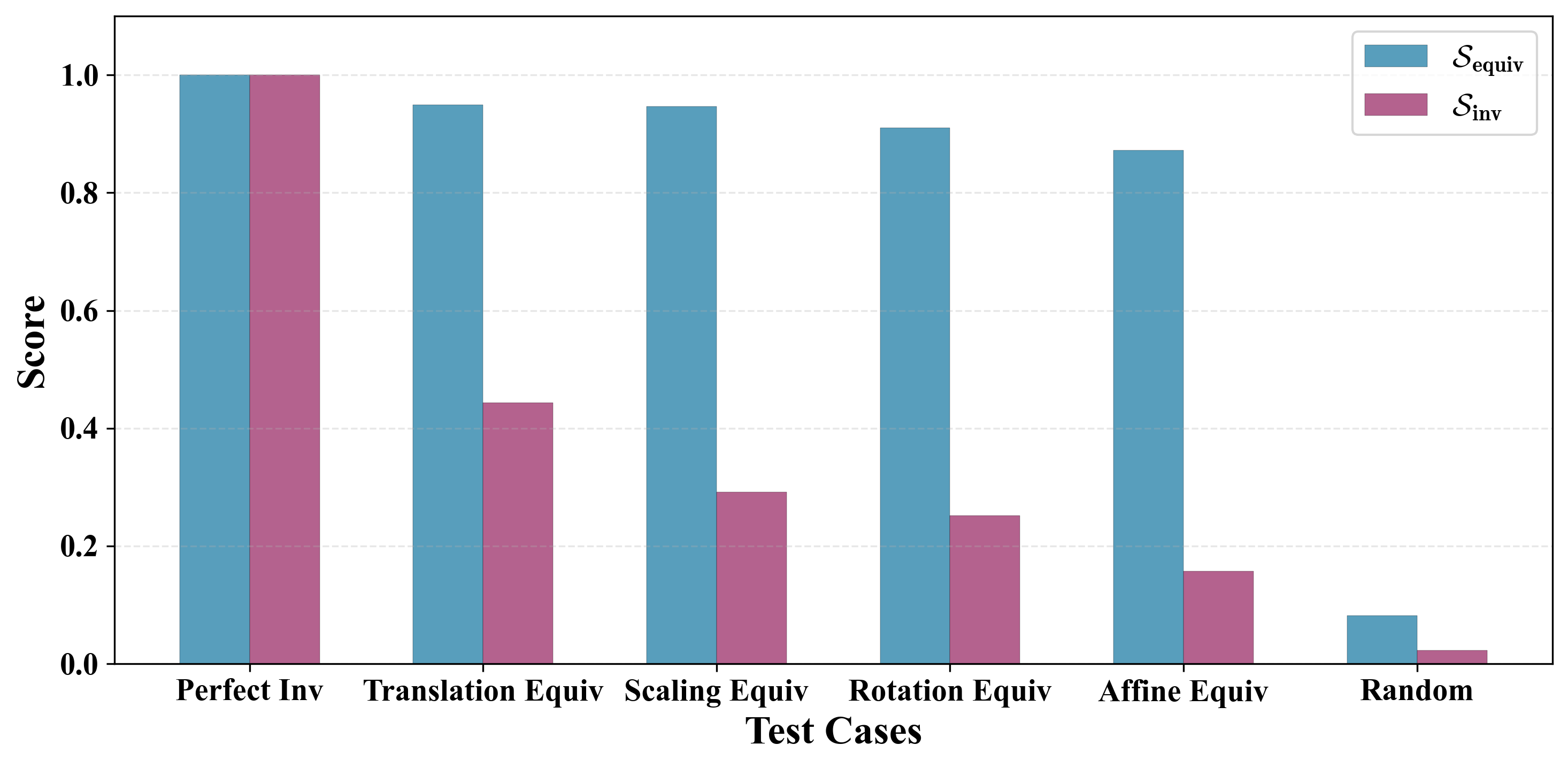}
    \caption{Validation of SEIS on MNIST activations using controlled transformations. The identity case (\textit{left}) yields high scores for both metrics, geometric transformations (\textit{middle}) preserve equivariance but reduce invariance, and the random baseline (\textit{right}) produces negligible scores.}
    \label{fig:exp1_svcca_validation}
\end{figure}

\noindent\textbf{Results.}
Figure~\ref{fig:exp1_svcca_validation} shows that SEIS reliably distinguishes between controlled regimes. In the identity condition, both scores approach $1.0$, confirming the correct calibration of the metrics. Under all geometric transformations, the invariance score drops sharply, indicating that the spatial feature basis changes. At the same time, the equivariance score remains high ($\mathcal{S}_\text{equiv} > 0.85$), correctly reflecting that the underlying information is preserved and remains linearly recoverable despite spatial redistribution. In contrast, the random baseline yields near-zero scores for both metrics, demonstrating that the high equivariance observed in the previous cases arises from structured geometric transformations rather than from spurious correlations in high-dimensional feature spaces.

\subsection{Depth-wise Evolution of Equivariance and Invariance in CNNs}
\label{sec:4-2}
\textbf{Experimental Setup.} We analyze how equivariance and invariance evolve across depth in standard convolutional networks during training by evaluating SEIS on internal representations induced by an input image and its affine transformed counterpart. We train ResNet architectures \cite{he2016deep} of varying scales on CIFAR-100 \cite{krizhevsky2009learning} for 200 epochs using SGD with momentum 0.9, weight decay $5e^{-4}$, and batch size 128. The initial learning rate is set to 0.1 and decayed by a factor of 0.2 at epochs 60, 120, and 160.

During evaluation, layer activations are computed on a fixed subset of the CIFAR-100 test set for each input image and for a transformed version obtained via a random affine transformation, with rotations in $[-15^\circ, 15^\circ]$, translations up to $5\%$ of the image size, and scaling in $[0.9, 1.1]$.

\begin{figure}[t]
    \centering
    \includegraphics[width=0.85\linewidth]{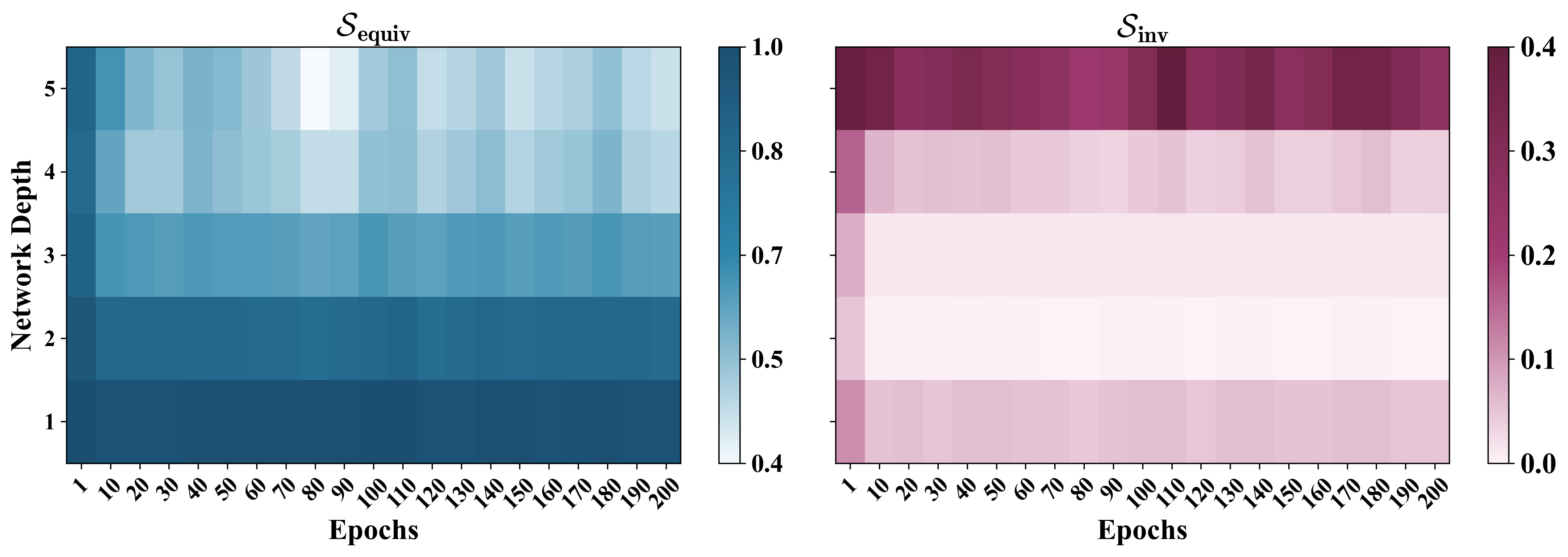}
    \caption{Depth-wise equivariance ($\mathcal{S}_\text{equiv}$) (\textit{left panel}) and invariance ($\mathcal{S}_\text{inv}$) (\textit{right panel}) scores across training epochs for ResNet-18 on CIFAR-100. Both properties stabilize early and exhibit a clear depth-dependent pattern.}
    \label{fig:exp2_training_dynamics_resnet18_cifar100}
\end{figure}

\noindent\textbf{Results.} Figure~\ref{fig:exp2_training_dynamics_resnet18_cifar100} shows the evolution of equivariance and invariance across network depth during standard training. Both scores stabilize within the first 10 epochs, indicating that these properties are established early and remain largely unchanged thereafter. A clear depth-dependent pattern emerges: equivariance is highest in early layers ($\mathcal{S}_\text{equiv}\approx0.95$) and gradually decreases with depth, while invariance exhibits the opposite trend, increasing from low values in shallow layers ($\mathcal{S}_\text{inv}\approx0.11$) to its highest value in the deepest layer ($\mathcal{S}_\text{inv}\approx0.34$). This progression is consistent with the hierarchical organization of classification networks, in which early layers preserve spatial structure and deeper layers develop more transformation-tolerant representations \cite{bau2017network}.

\subsection{Factors Shaping Equivariance and Invariance}
\subsubsection{Effect of Data Augmentation}
\label{sec:4-3-1}
\begin{wrapfigure}[13]{r}{0.45\textwidth}
    \centering
    \vspace{-0.15in}
    \includegraphics[width=0.43\textwidth]{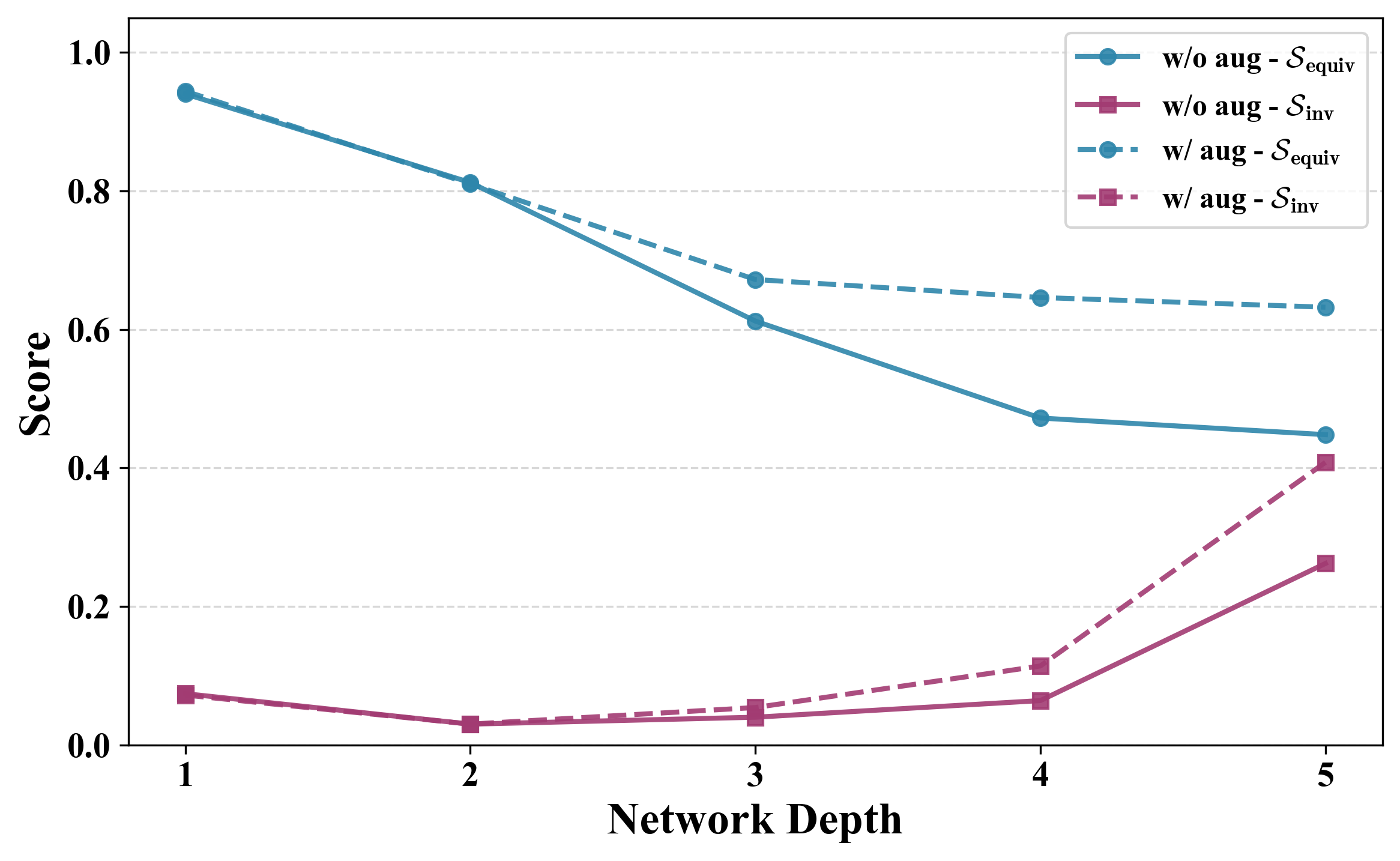}
    \caption{$S_{\text{equiv}}$ and $S_{\text{inv}}$ across network depth for ResNet-18 on CIFAR-100 classification.}
    \vspace{-2pt}
    \label{fig:exp2_data_aug_resnet18_cifar100}
\end{wrapfigure}
Data augmentation is a primary mechanism for encouraging invariance in deep networks~\cite{lyle2020benefits}. We examine the effect of training-time data augmentation by comparing models trained with (using the same family of affine transformations during evaluation in Section~\ref{sec:4-2}) and without affine augmentations.

For classification, we train ResNets on CIFAR-100 under identical optimization settings in Section~\ref{sec:4-2}, differing only in the use of affine augmentations. We observe qualitatively similar trends across ResNet architectures of varying depth (see Appendix~\ref{sec:appendix_resnet_variants}). As shown in Figure~\ref{fig:exp2_data_aug_resnet18_cifar100} for ResNet-18, augmented models maintain higher equivariance in deeper layers ($\mathcal{S}_\text{equiv}\approx0.62$ versus $\approx0.45$ without augmentation), indicating improved preservation of geometric information. At the same time, invariance is consistently strengthened across depth, with the deepest layer reaching $\mathcal{S}_\text{inv}\approx0.41$ compared to $\approx0.28$ in the non-augmented model. These results indicate that affine augmentation promotes representations that are more transformation-tolerant while preserving equivariant structure, consistent with empirical observations based on Pearson correlation measures~\cite{bruintjes2023affects}.

\begin{wrapfigure}[13]{r}{0.45\textwidth}
    \centering
    \includegraphics[width=0.43\textwidth]{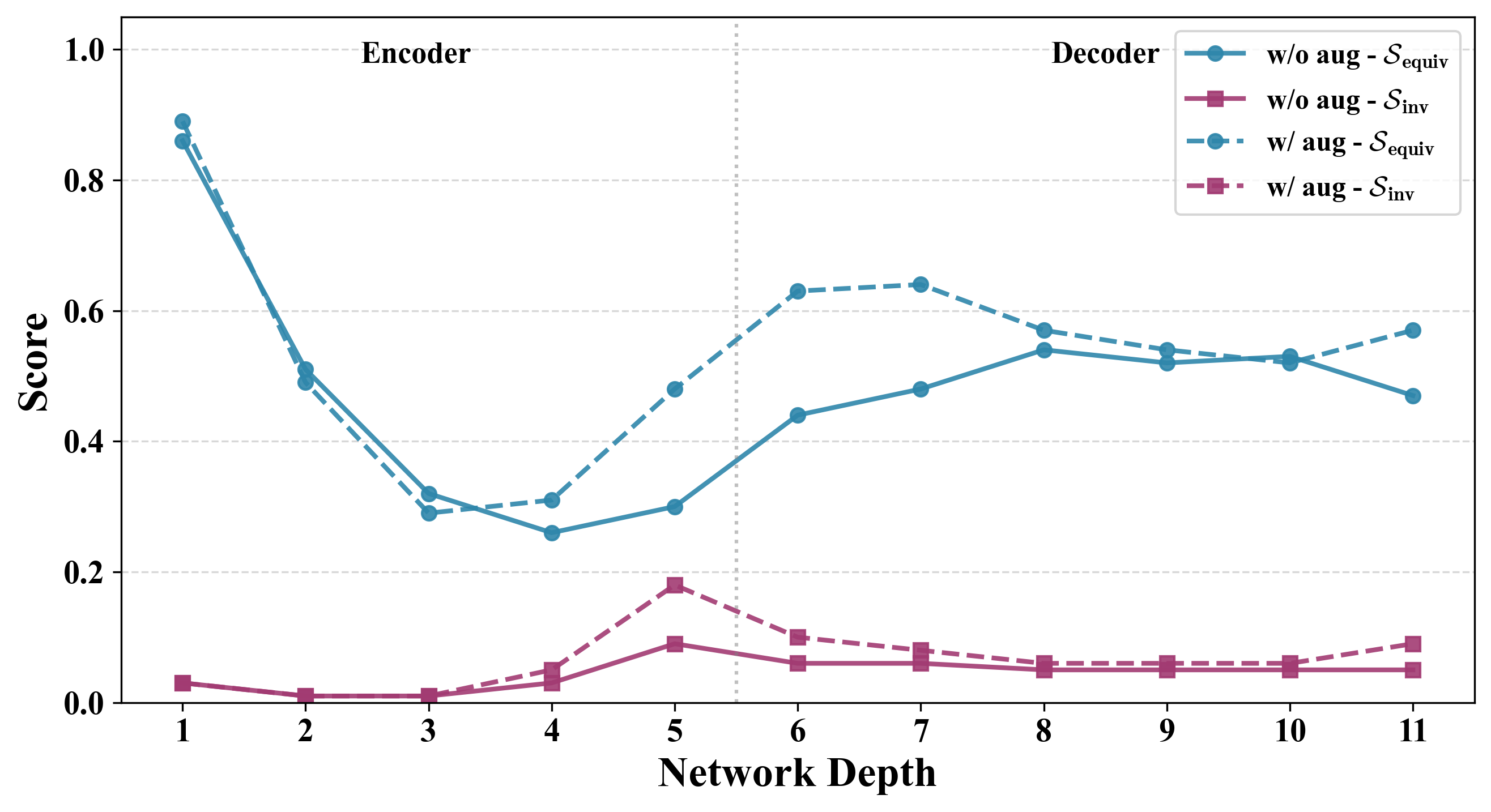}
    \caption{$S_{\text{equiv}}$ and $S_{\text{inv}}$ across network depth for ResNet-FCN on PASCAL VOC 2012 segmentation.}
    \label{fig:exp2_data_aug_resfcn_pascal}
\end{wrapfigure}

We further evaluate this effect in a segmentation setting. Specifically, we use a ResNet-based encoder with a Fully Convolutional Network decoder \cite{long2015fully} trained on PASCAL VOC 2012 \cite{everingham2011pascal}, again comparing models trained with and without affine augmentation under identical settings. Figure~\ref{fig:exp2_data_aug_resfcn_pascal} shows that affine augmentation has a distinct effect compared to classification.
In the encoder, the depth-wise trends of equivariance and invariance closely mirror those observed in classification: equivariance decreases while invariance increases toward deeper layers, and augmentation strengthens both properties, particularly at the bottleneck (e.g., $\mathcal{S}_\text{inv}\approx0.18$ vs.\ $\approx0.09$). In contrast, the decoder exhibits different behavior. Augmentation significantly increases equivariance in early decoder layers (e.g., $\mathcal{S}_\text{equiv}\approx0.63$ vs.\ $\approx0.44$ at layer 6), indicating improved recovery of spatial structure during upsampling, while invariance remains relatively low throughout. 

These results show that the effect of augmentation is consistent with classification in the encoder, but diverges in the decoder, where it primarily enhances equivariance in layers responsible for spatial reconstruction.

\begin{figure}[b]
\centering
\subfloat{\includegraphics[width=0.41\linewidth]{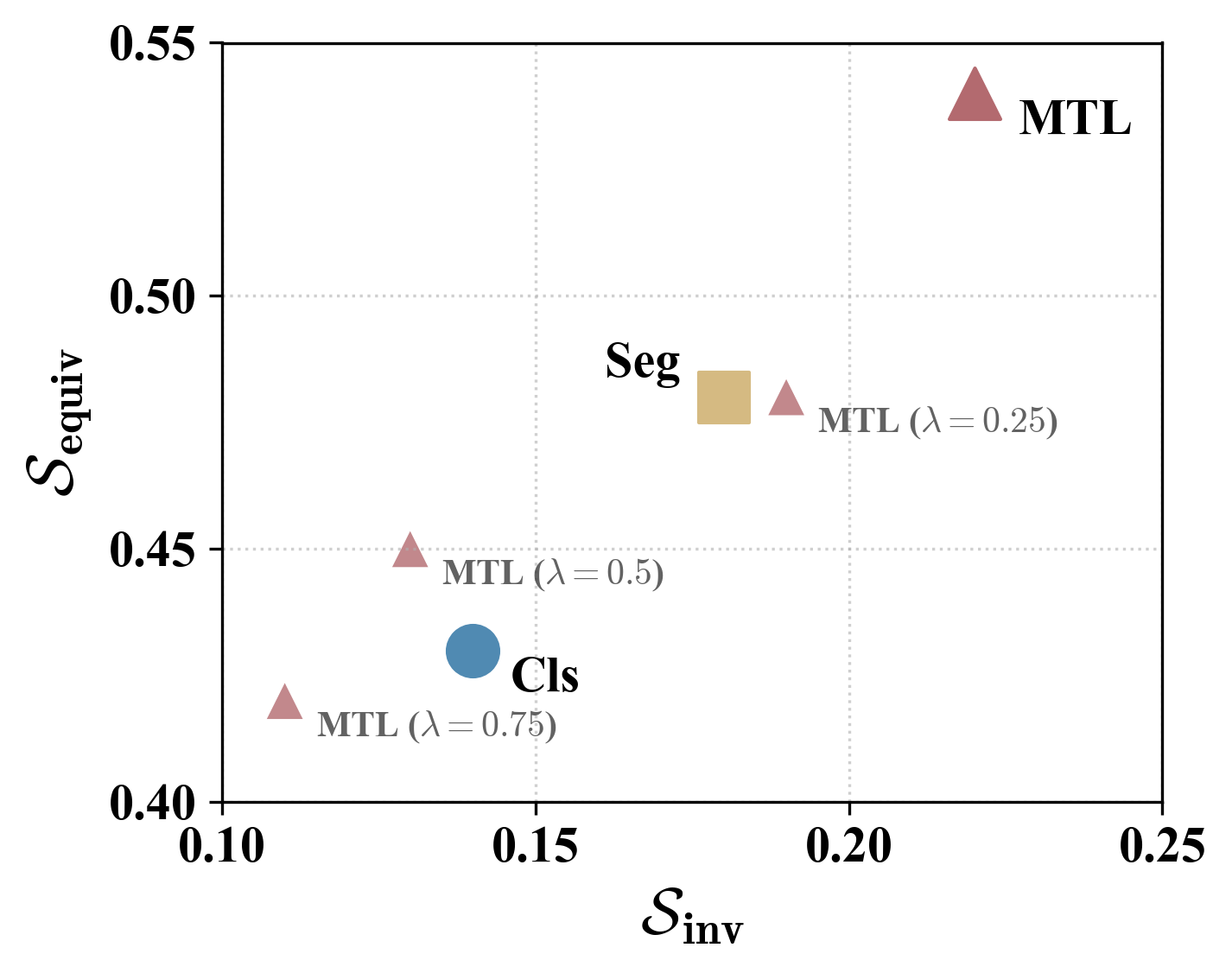}}
\hfil
\subfloat{\includegraphics[width=0.5\linewidth]{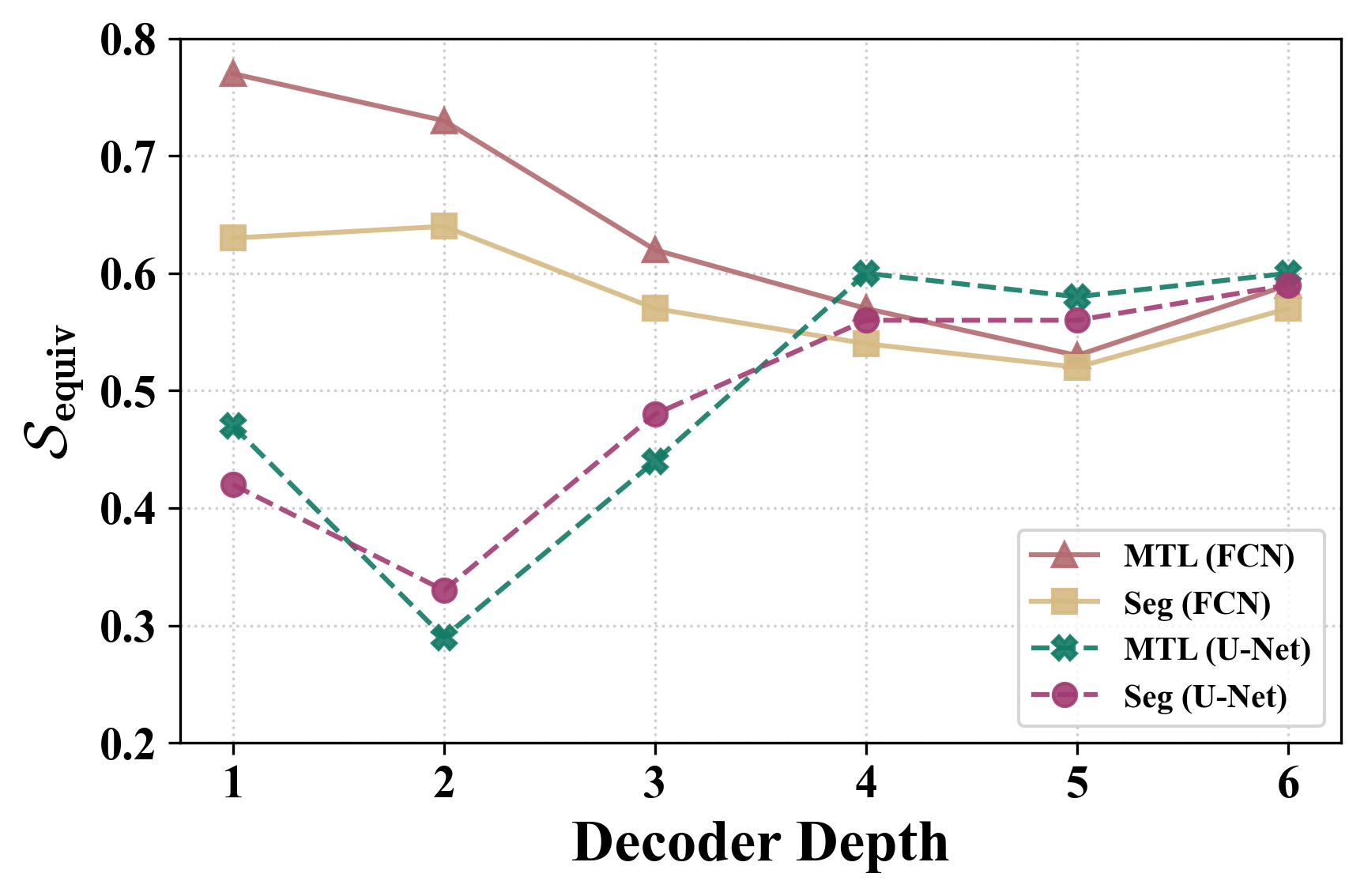}}
\caption{\textit{Left panel:}  Equivariance ($\mathcal{S}_\text{equiv}$) versus invariance ($\mathcal{S}_\text{inv}$) at the bottleneck layer for Classification (Cls), Segmentation (Seg), and MTL (Multi-Task Learning). \textit{Right panel:} Equivariance trajectories across decoder depth, comparing FCN and U-Net decoders under single-task (Seg) and multi-task objectives.}
\label{fig:task_objs}
\end{figure}

\subsubsection{Effect of Task Objectives}
Different computer vision tasks impose distinct geometric requirements on learned representations. Image segmentation requires precise spatial localization and therefore benefits from equivariance \cite{chidester2019enhanced}, whereas image classification prioritizes abstraction over spatial transformations and thus benefits from invariance \cite{zhu2021understanding}. We investigate whether these task-specific objectives systematically shape the geometric properties encoded in neural representations, and whether Multi-Task Learning (MTL) \cite{caruana1997multitask} induces a predictable trade-off between equivariance and invariance in shared encoder layers.

\noindent\textbf{Experimental Setup.} We train three model variants on PASCAL VOC 2012: (i) classification-only using image-level labels, (ii) segmentation-only using pixel-level masks, and (iii) MTL with a shared encoder and task-specific decoders. All models employ an identical ResNet-50 backbone as the encoder with an FCN decoder, and are optimized using SGD with momentum 0.9. The initial learning rate is set to 0.1 and decayed by a factor of 0.2 at epochs 60 and 120, with a total training duration of 200 epochs. For the MTL setting, we use a weighted objective $\mathcal{L}_\text{seg} + \lambda \mathcal{L}_\text{cls}$ with $\lambda \in \{0.25, 0.50, 0.75, 1.0\}$ to examine how task balance influences the equivariance--invariance trade-off. Equivariance and invariance are quantified using SEIS at each layer under random affine transformations (the same family of transformations used in Section \ref{sec:4-2}) applied to a held-out test set.

\noindent\textbf{Results.} Figure~\ref{fig:task_objs} \textit{left} shows that the geometric properties of the bottleneck layer diverge significantly depending on the task objective. Surprisingly, the MTL model ($\lambda=1.0$) achieves both the highest invariance ($\mathcal{S}_\text{inv}\approx0.22$) and the highest equivariance ($\mathcal{S}_\text{equiv}\approx0.54$) among all models. This supports the hypothesis that complementary objectives act as an inductive bias, encouraging the shared encoder to balance spatial precision with semantic abstraction \cite{vandenhende2021multi}. In contrast, the classification model exhibits the lowest scores, attributed to global average pooling, which removes the penalty for spatial misalignment. Notably, this synergy is sensitive to task balance; reducing the classification weight $\lambda$ progressively erodes these structural gains, shifting representations toward single-task regimes.

\noindent\textbf{Decoder Architecture Matters.}
We further examine how architectural design interacts with task objectives by comparing two segmentation decoders: an FCN and a U-Net \cite{ronneberger2015u}. These architectures differ primarily in their use of skip connections, allowing us to isolate their role in preserving spatial information. Figure~\ref{fig:task_objs} \textit{right} shows that the FCN decoder exhibits an obvious decrease in equivariance across depth, reflecting its reliance on progressively abstracted features from the encoder. Although MTL-FCN starts with higher equivariance (0.77) than the single-task segmentation baseline (0.63), this advantage diminishes as depth increases. In contrast, the U-Net decoder demonstrates a non-monotonic equivariance profile, with an initial decrease followed by a pronounced recovery in deeper layers. This behavior indicates that skip connections act as a structural bypass, reintroducing high-resolution spatial features from earlier encoder stages and restoring equivariance that would otherwise be lost during decoding. These findings highlight the critical role of architectural mechanisms in modulating the propagation of geometric structure through the network.

\begin{figure}[t]
    \centering
    \includegraphics[width=\linewidth]{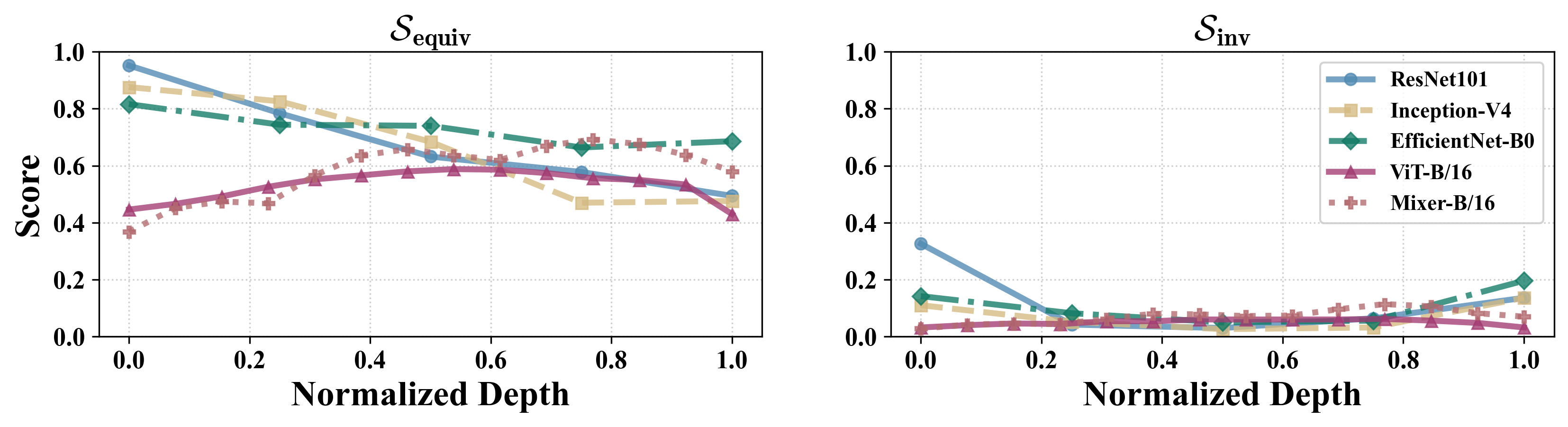}
    \caption{SEIS across architectures. Equivariance ($\mathcal{S}_\text{equiv}$) and invariance ($\mathcal{S}_\text{inv}$) are plotted over normalized depth for ImageNet-pretrained models evaluated on Imagenette. CNNs exhibit a consistent decrease in equivariance alongside a late emergence of invariance, while ViTs show weaker invariance and a distinct equivariance trajectory, and MLP-Mixers display intermediate behavior.}
    \label{fig:exp4_archs_comparison}
\end{figure}

\subsection{Equivariance and Invariance Across Architectures}
\subsubsection{CNNs, ViTs, and MLP-Mixers}
We next examine whether the depth-wise behavior observed in convolutional networks holds across different model designs. In particular, we compare convolutional models with architectures that process images as sequences, which do not explicitly encode spatial structure through convolution.

\noindent\textbf{Experimental Setup.}
We evaluate a range of ImageNet-pretrained \cite{deng2009imagenet} models on the Imagenette dataset \cite{Howard_Imagenette_2019}, including ResNet-101 \cite{he2016deep}, Inception-V4 \cite{szegedy2017inception}, EfficientNet-B0 \cite{tan2019efficientnet}, ViT-B/16 \cite{dosovitskiy2021image}, and Mixer-B/16 \cite{tolstikhin2021mlp}. For each model, we apply random affine transformations (using the same family of transformations as in Section~\ref{sec:4-2}) to input images and compute SEIS scores between the activations of the original and transformed inputs at multiple depths.

For convolutional networks, activations have shape $(b, c, h, w)$ and are reshaped into matrices over spatial locations as described in Section~\ref{sec:method}. For models that process images as sequences, activations are given as $(b, n, d)$, where $n$ denotes the number of patches and $d$ the feature (embedding) dimension, analogous to the channel dimension $c$ in convolutional networks. In this case, we treat each patch as a spatial unit and reshape the activations into $(n, b \cdot d)$ matrices. The class token is excluded from this analysis, as it does not correspond to a specific image region.

\noindent\textbf{Results.}
Figure~\ref{fig:exp4_archs_comparison} shows the evolution of equivariance and invariance of these different architecture designs across depth. Convolutional models exhibit a consistent pattern: equivariance decreases with depth, while invariance emerges in later layers. For example, in ResNet-101, equivariance drops from $0.94$ to $0.51$, while invariance is suppressed in intermediate layers and recovers to $0.14$ at the output. Inception-V4 follows a similar trend. EfficientNet-B0 maintains relatively high equivariance throughout and shows the strongest late-stage invariance among the convolutional models, reaching $0.28$. These results indicate that convolutional architectures progressively shift from spatially structured responses in early layers to more transformation-tolerant representations in deeper layers, with variations in strength across designs.

In contrast, the transformer model behaves differently. Equivariance starts at $0.44$, increases to approximately $0.59$ in intermediate layers, and decreases to $0.47$ at the output. Invariance remains low throughout, typically between $0.03$ and $0.06$, indicating limited development of transformation-invariant features under the same conditions.

The MLP-Mixer displays an intermediate pattern. Equivariance increases steadily from $0.36$ to around $0.60$, while invariance increases moderately from $0.03$ to $0.12$ before stabilizing near $0.09$. This behavior differs from both convolutional networks and the transformer model, suggesting that its combination of patch-wise and channel-wise processing leads to a distinct balance between sensitivity and robustness to transformations.

\subsubsection{Role of Positional Encoding in ViTs}
We examine how positional encoding influences equivariance and invariance in ViTs. We use a ViT model pretrained on ImageNet and evaluate it on the Imagenette dataset. Positional embeddings are modified only at inference time, without any additional training. We consider three variants in which positional information is removed, shuffled, or replaced with random noise, enabling a controlled comparison that isolates the contribution of positional encoding to the learned representations.

\begin{wraptable}[10]{r}{0.45\textwidth}
\centering
\vspace{-0.05in}
\begin{tabular}{lcc}
\toprule
Model & Mean $S_{\text{equiv}}$ & Mean $S_{\text{inv}}$ \\
\midrule
ViT (normal)   & 0.56 & 0.05 \\
ViT (no pos)   & 0.47 & 0.04 \\
ViT (shuffle)  & 0.46 & 0.05 \\
ViT (random)   & 0.45 & 0.04 \\
\bottomrule
\end{tabular}
\caption{Average equivariance ($\mathcal{S}_\text{equiv}$) and invariance ($\mathcal{S}_\text{inv}$) for ViT positional variants.}
\label{tab:vit_pos}
\end{wraptable}

Table~\ref{tab:vit_pos} summarizes the mean SEIS across layers. The standard model achieves the highest values, with mean equivariance of $0.56$ and mean invariance of $0.05$. In contrast, removing or corrupting positional encoding consistently reduces equivariance to approximately $0.45$--$0.47$, while invariance remains similar ($\approx 0.04$--$0.05$) across all variants.

This indicates that positional encoding improves consistency under transformations, while its absence or corruption leads to uniformly weaker responses. The similarity among these variants suggests that once spatial correspondence is disrupted, the specific form of corruption has little additional effect.

\section{Conclusion}
We introduced SEIS, a subspace-based metric that quantifies how geometric information is preserved in neural representations by separating equivariance from invariance without requiring explicit knowledge of the transformation. By framing geometric consistency as subspace alignment, SEIS provides an interpretable characterization of how spatial information is preserved or re-encoded across layers. Across a wide range of architectures and tasks, we observe consistent yet non-trivial patterns: convolutional encoders transition from equivariance to invariance with depth, while segmentation decoders partially reverse this trend. Crucially, data augmentation and multi-task learning strengthening both properties simultaneously, challenging the conventional view of a strict trade-off between the two. Extending beyond CNNs, transformer-based and MLP-based models exhibit distinct geometric trajectories, highlighting the role of architectural design in shaping representation structure.

\section*{Limitations and Potential Directions}
SEIS measures equivariance and invariance through linear subspace alignment, which may not fully capture non-linear relationships in representations. The method also treats spatial locations or patches as comparable units across transformations, which may be less well-defined for architectures without explicit spatial structure. Future work could extend SEIS to incorporate non-linear similarity measures (e.g., kernel-based methods), explore a broader range of transformations, and develop more principled ways to compare representations across architectures.

\newpage



\bibliographystyle{plainnat}
\bibliography{refs}


\appendix
\section{Depth-wise Evolution in Deeper ResNet Architectures}
\label{sec:appendix_resnet_variants}
\begin{figure}[!htbp]
    \centering
    \subfloat[ResNet-50]{\includegraphics[width=0.48\linewidth]{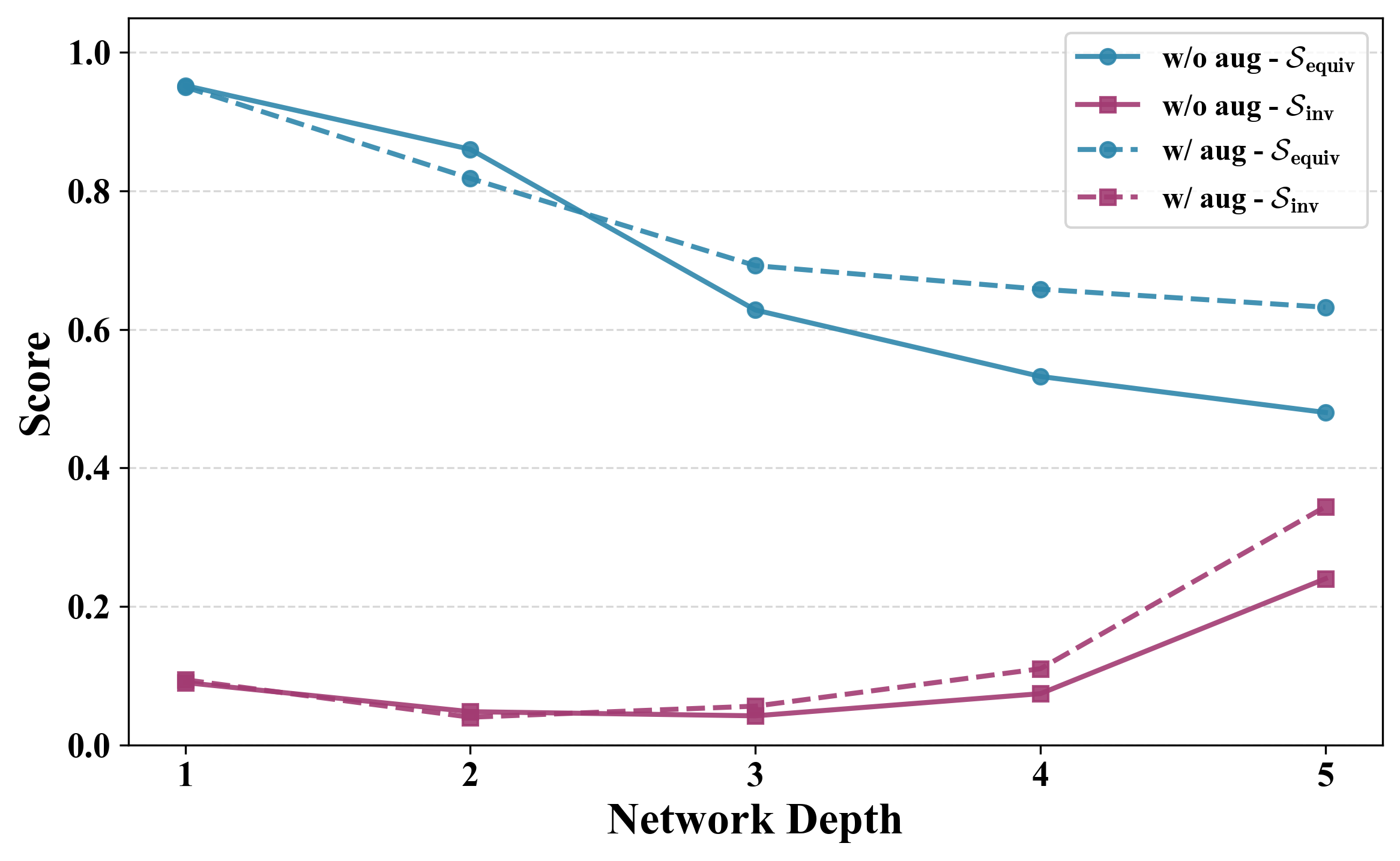}}
    \hfill
    \subfloat[ResNet-101]{\includegraphics[width=0.48\linewidth]{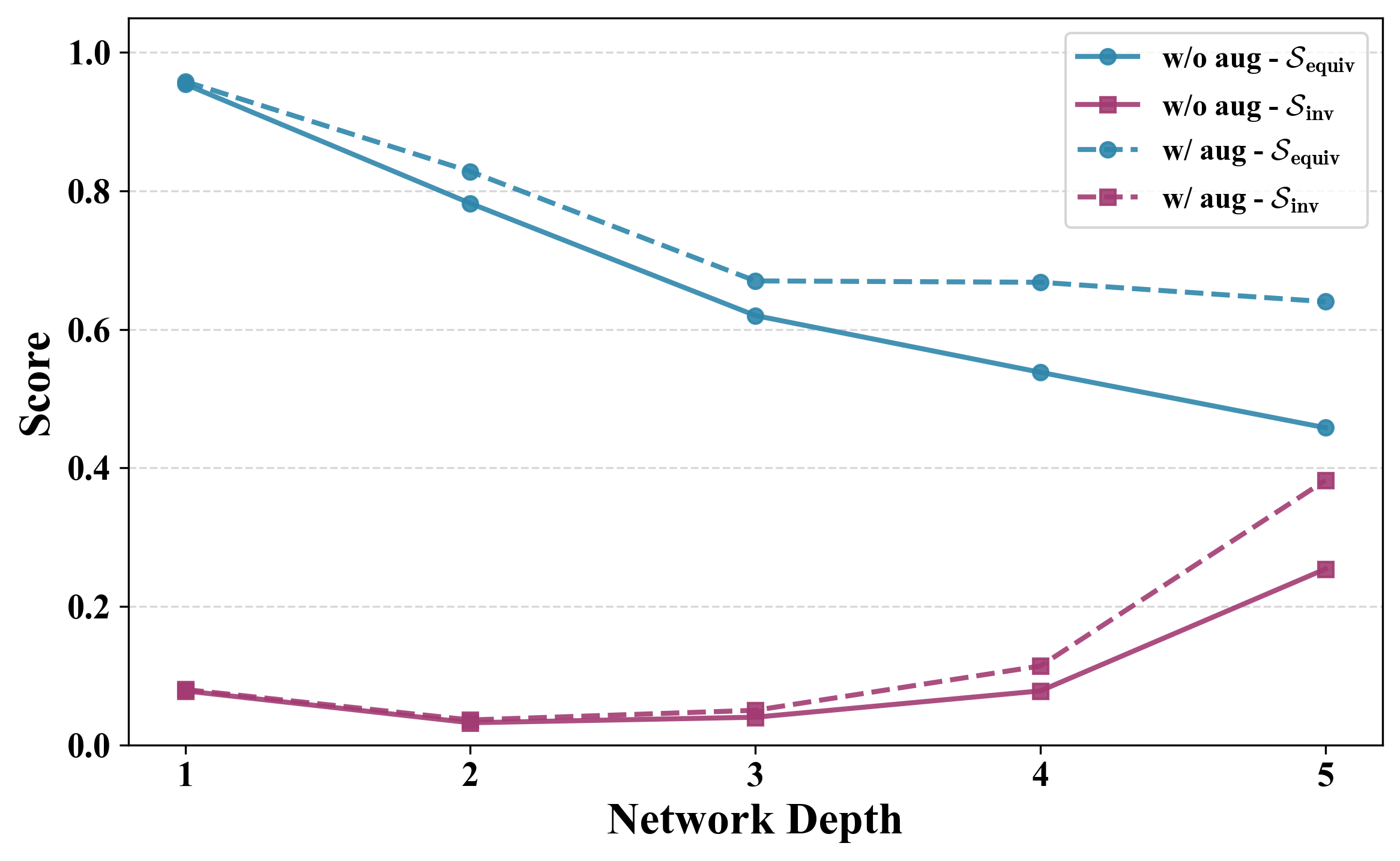}}
    \caption{Evolution of equivariance and invariance across network depth for deeper ResNet variants on CIFAR-100. Both models exhibit consistent depth-wise trends, with decreasing equivariance and increasing invariance toward deeper layers.}
    \label{fig:appendix_resnet_variants}
\end{figure}

To verify that the observed depth-wise behavior is consistent across model scales, we report additional results for ResNet-50 and ResNet-101 trained under the same setting as described in Section~\ref{sec:4-3-1}. Both models are trained on CIFAR-100 for 200 epochs with identical optimization and augmentation configurations. As shown in Figure~\ref{fig:appendix_resnet_variants}, both architectures exhibit trends consistent with ResNet-18: equivariance decreases with depth while invariance increases in deeper layers. This confirms that the equivariance--invariance transition is a stable property across ResNet architectures of varying depth. Furthermore, the effect of data augmentation observed in ResNet-18 generalizes consistently across scales, where augmented models maintain higher equivariance in deeper layers and stronger invariance throughout.

\section{Variance Threshold Sensitivity}
\label{sec:appendix_vt_sensitivity}
\begin{figure}[!htbp]
    \centering
    \includegraphics[width=\linewidth]{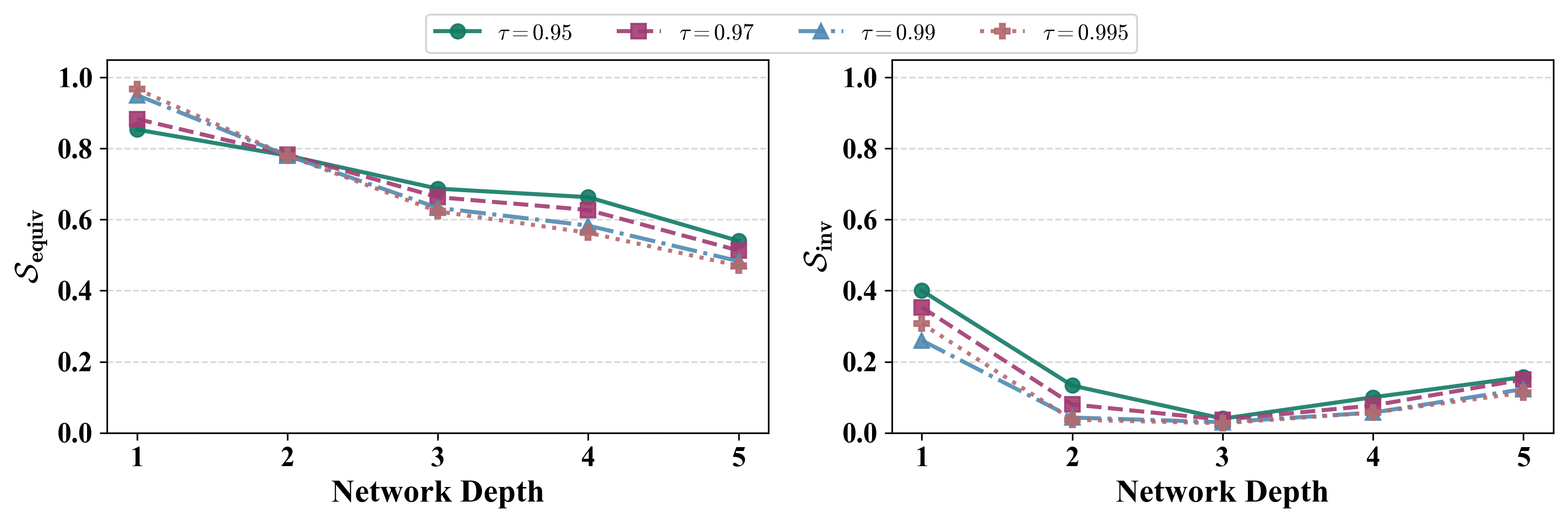}
    \caption{Sensitivity of SEIS to the SVD variance threshold. Equivariance ($\mathcal{S}_\text{equiv}$, left) and invariance ($\mathcal{S}_\text{inv}$, right) scores across network depth for an ImageNet-pretrained ResNet-101 evaluated on Imagenette, computed under four variance thresholds used in the SVD truncation step (Section~\ref{sec:3-3}).}
    \label{fig:appendix_vt_sensitivity}
\end{figure}

We evaluate the sensitivity of SEIS to the variance threshold hyperparameter governing SVD truncation, varying it across \{0.95, 0.97, 0.99, 0.995\} on an ImageNet-pretrained ResNet-101 evaluated on Imagenette. As shown in Figure~\ref{fig:appendix_vt_sensitivity}, the depth-wise trends of both $\mathcal{S}_\text{equiv}$ and $\mathcal{S}_\text{inv}$, remain qualitatively consistent across all tested thresholds: equivariance decreases monotonically with depth while invariance remains low in intermediate stages before recovering slightly at the output. These results confirm that the qualitative behavior of SEIS is robust to the choice of variance threshold, and that the default value of 0.99, following standard practice in SVCCA \cite{raghu2017svcca}, provides a stable and representative operating point.

\end{document}